\pdfoutput=1

\documentclass[11pt]{article}

\usepackage[final]{coling}

\usepackage{times}
\usepackage{latexsym}

\usepackage[T1]{fontenc}

\usepackage[utf8]{inputenc}

\usepackage{microtype}

\usepackage{inconsolata}

\usepackage{graphicx}
\usepackage{pdfpages}

\usepackage{arabtex}
\usepackage{utf8}
\usepackage[T1]{fontenc}
\newcommand{\hide}[1]{}

\newcommand{\AHAMZAUP}{{\^{A}}}

\newcommand{\AHAMZADN}{{\v{A}}}

\newcommand{\TAMARBUTA}{{$\hbar$}}

\newcommand{\SHIN}{{\v{s}}}
\newcommand{\AYN}{{$\varsigma$}}

\newcommand{\SHADDA}{{$\sim$}}

\newcommand{\chainbank}{{\textsc{ChainBank}}}
\newcommand{\camelmorph}{{\textsc{CamelMorph}}}

\title{A Derivational Chain Bank for Modern Standard Arabic}

\author{
 \textbf{Reham Marzouk,\textsuperscript{1,2}}
 \textbf{Sondos Krouna,\textsuperscript{3}}
\textbf{Nizar Habash\textsuperscript{1}}
\\
\\
 \textsuperscript{1}Computational Approaches to Modeling Language (CAMeL) Lab,\\ New York University Abu Dhabi\\ 
 \textsuperscript{2}Information Technology Department, IGSR, Alexandria University\\
 \textsuperscript{3}ISLT, Carthage University
 \\
 {\texttt{
 \href{mailto:email@domain}{marzoukreham@gmail.com, sondes.krouna@islt.ucar.tn, nizar.habash@nyu.edu
}}
 }
}

\begin{document}
\setcode{utf8}
\maketitle
\begin{abstract}
We introduce the new concept of an \textit{Arabic Derivational Chain Bank} ({\chainbank}) to leverage the relationship between form and meaning in modeling Arabic derivational morphology. We constructed a knowledge graph network of abstract patterns and their derivational relations, and aligned it with the lemmas of the {\camelmorph} morphological analyzer database. This process produced chains of derived words' lemmas linked to their corresponding lemma bases through derivational relations, encompassing 23,333  derivational connections. 
The {\chainbank} is publicly available.\footnote{\url{https://github.com/CAMeL-Lab/ArabicChainBank}\label{github}}

\end{list} 
\end{abstract}


\section{Introduction}
Lexical resources are essential for improving the accuracy of language processing and pedagogical applications, as they enhance computational systems' ability to grasp the nuanced meanings and contextual variations of human language. Despite significant efforts, the Arabic language still lacks tools that focus on its compositional morphological structure and semantic connections. 
Derivational modeling offers a computational framework to capture the interplay between word form and meaning, clarifying Arabic's complex derivational pathways and resolving its structural ambiguities.

Arabic derivational morphology is fundamentally tied to its templatic system, where roots and patterns provide different types of semantic abstractions to express multiple meanings \cite{Gadalla:2000:comparative,holes2004modern,Habash:2010:introduction}. 
%
The process of deriving words from roots is not consistent, leading to challenges that hinder the understanding of the meanings of derived words and pose significant obstacles for derivational modeling. For instance, a single pattern can convey different derivational meanings, resulting in ambiguity among derived words that share the same root. As an example, the \textit{masdar/verbal noun} \<المصدر> and the \textit{descriptive adjective} \<الصفة المشبهة>  may share the same pattern \textit{1a2A3}, e.g. \<حصاد> \textit{HaSAd}\footnote{\newcite{Habash:2007:arabic-transliteration}'s Arabic transliteration scheme.} `harvest' and the adjective \<جبان> \textit{jabAn} `coward'. Likewise, Homographs can be derived from the same base to convey distinct meanings; consequently, each word possesses a different set of derivatives.
%
For example, each of the two senses of the verb \<فلح> \textit{falaH} `to succeed' and `to farm' has its own masdar:  \<فلاح> \textit{falAH} `success' and \<فلاحة> \textit{filAHa{\TAMARBUTA}} `farming'.
Another crucial behavior is the meaning shift of some derivatives from the original abstract meaning of the root,
e.g., \<كتيبة> \textit{katiyba{\TAMARBUTA}} `battalion' is ultimately derived from the root
\<ب>.\<ت>.\<ك>~\textit{k.t.b} `writing-related'. 

Interpreting the behavior of derived words in the Arabic language, along with the deviations from derivational rules, necessitates a robust organization of derivatives within a framework capable of tracing the various paths of derivation and managing the resulting ambiguities. The objective of this study is to define the new concept of the \textit{Arabic Derivational Chain Bank} (henceforth, {\chainbank}), which serves as the first representation of the Arabic derivational structure. The {\chainbank} presents connected chains that illustrate the path of each derived word and the relation between connected words by providing their derivational meanings. To construct the {\chainbank}, we employed a knowledge graph structure to build a network of abstract patterns, along with a classification model to align this network with lexical database of the Arabic morphological analyzer the {\camelmorph}  \cite{khairallah-etal-2024-camel} based on selected features. 
The {\chainbank} is a morphological model that exploits Arabic's compositional morphological and semantic features while accommodating ad hoc exceptions.

\section{Related Work}

\paragraph{Computational Derivational Morphology}
Several studies have modeled derivational morphology using a range of techniques.
%
\newcite{habash-dorr-2003-catvar} clustered categorial variations of English lexemes to develop the CATVAR resource. Similarly, \newcite{zeller2013derivbase} created DERIVBASE, a derivational resource for German, using a rule-based framework to induce derivational families (i.e., clusters of lemmas in derivational relationships).  \newcite{hathout-namer-2014-demonette} developed Démonette by integrating two lexical resources and applying rules to link words to their bases while considering their semantic types. Following Zeller’s approach, \newcite{vodolazsky2020derivbase} and \newcite{snajder-2014-derivbase} constructed derivational models for Russian and Croatian, respectively. \newcite{kanuparthi2012hindi} introduced a derivational morphological analyzer for Hindi built on a mapping from an inflectional analyzer.
\newcite{cotterell2017paradigm} argued for a paradigmatic treatment of
derivational morphology and used sequence-to-sequence models to learn mappings from fixed paradigm slots to their corresponding derived forms. \newcite{hofmann-etal-2020-graph} proposed a 
graph auto-encoder that learns embeddings capturing information about the compatibility of affixes and stems in derivation.

\paragraph{Arabic Computational Morphology}
Research on Arabic computational morphology has primarily focused on inflectional modeling \cite{Kiraz:1994:multi-tape,Beesley:1998:arabic,Al-Sughaiyer:2004:arabic,Habash:2006:magead,Taji:2018:arabic}. This focus has led to the development of various models for morphological analysis, generation, and disambiguation.  \newcite{Habash:2012:mada+tokan} introduced MADA, a tool designed to analyze and disambiguate Arabic morphology in context. \newcite{Pasha:2014:madamira} developed MADAMIRA, which identifies the morphological features of a word and ranks analysis results based on their compatibility with the model's predictions. More recently, tools such as CALIMA-Star \cite{Taji:2018:arabic} and {\camelmorph} \cite{habash2022morphotactic,khairallah-etal-2024-computational,khairallah-etal-2024-camel} have emerged as advanced morphological analyzers and generators, with a wide range of features.
A few efforts have incorporated derivational features to enhance their models. For instance, the morphological analyzer Al Khalil Morph system \cite{boudlal2010alkhalil,boudchiche2017alkhalil} utilizes a database categorized into derived and non-derived classes based on root, vocalized, and unvocalized patterns. Additionaly, \newcite{zaghouani2016ampn} conducted a pilot study aimed at representing the derivational structure of roots and patterns while addressing the multiple senses associated with a single pattern. However, none of these studies developed a  comprehensive model focused extensively on  derivational morphology.

Inspired by the efforts on systematic treatment of derivational morphology in other languages, we propose a model that captures the complexity and elegance of the Arabic derivational system.


\section{Arabic Derivational Morphology Terms}
\label{terms}

In Arabic templatic morphology, discontinuous consonantal morphemes, \textbf{roots}, interconnect with different  \textbf{patterns} of vowels and consonants to construct different meanings.  
Each root has a general semantic meaning and each pattern is associated with a certain \textbf{canonical meaning}. 
The set of words sharing the same root, a \textbf{derivational family}, are organizable as a derivational network connecting  hierarchically up to a (typically) single base word. 
%
%
%
Derived words can be either \textbf{canonical}, where the word's meaning matches its pattern's meaning, or \textbf{non-canonical}, where an ad hoc deviation of regular form occurs.
For example the two words
\<ضرب> \textit{Darb} `hitting', and 
\<شمس> \textit{{\SHIN}ams} `sun', share the same pattern \textit{1a23}, whose canonical meaning is the masdar, matching the former (canonical) but not the latter (non-canonical).
Derived words can also be formed with  \textbf{derivational affixes}, e.g.,  the suffix \<ي>+ \textit{+iy{\SHADDA}} (\<ياء النسبة>  {Attributive yA'}) appends to the base \<علم> \textit{{\AYN}ilm} `science' to produce the attributive adjectives  \<علمي> \textit{{\AYN}ilmiy{\SHADDA}} `scientific'.
%
%
Verbs are divided into \textbf{unaugmented}, which are composed of roots and vocalism-only patterns, and \textbf{augmented}, which are derived from unaugmented verbs by geminating, lengthening of vowels, prefixation or infixation \cite{Gadalla:2000:comparative}. 
Nouns are categorized into \textbf{primary nouns}, which are directly derived from roots \cite{Gadalla:2000:comparative}, and \textbf{derived nouns}, which originate from verbs and encompass derivational classes such as verbal nouns (masdar), nouns of location, etc. 
In some cases, derived words involve shifting the meaning to a contextually unrelated interpretation of their base form, i.e., \textbf{semantic specification}. For instance, the noun \<مكتوب> \textit{maktuwb} `letter/message' is derived from the passive participle \<مكتوب>  \textit{maktuwb} `written".


\section{The \textsc{ChainBank} Framework}
The role of the Arabic derivational {\chainbank} is to systematically link all derivatives belonging to the same derivational family in a sequential manner (chain), starting from the root and progressing through each derived form. This chain establishes a clear relationship between each derivative and its base, clarifying the morphological processes that generate new words. By organizing derivatives in this structured form, the derivational chain highlights the hierarchical and interdependent nature of word formation, providing insights into how base forms evolve into more complex derivatives while preserving their semantic and grammatical connections.

We represent the {\chainbank} as a dynamic tree-structured knowledge graph starting with the root. Each node in this graph corresponds to a derived word and includes its morphosemantic attributes, such as pattern, part-of-speech, functional features, and lexical meaning. The connections between pairs of nodes denote the derivational relationships of each child node to its base parent.

To create the {\chainbank}, we developed an extensive network that represents the organization of abstract patterns, such as \<فَعَل>  
\textit{CaCaC/1a2a3}
and \<فَعِيل>
\textit{CaCiyC/1a2iy3}, and integrated this network with the {\camelmorph} lexical database. This combination forms a large-scale network connecting Arabic words through their derivational relationships. 
The process includes two levels: 
\begin{itemize}
    \item The \textbf{abstract level} focuses on the abstract patterns designed to represent various derivatives.
 \item The \textbf{concrete level} is where abstract patterns are linked to lemmas to produce derived words along with their derivational meanings.  
\end{itemize}

\begin{figure*}[ht]
    \begin{center}
        \includegraphics[width=\textwidth]{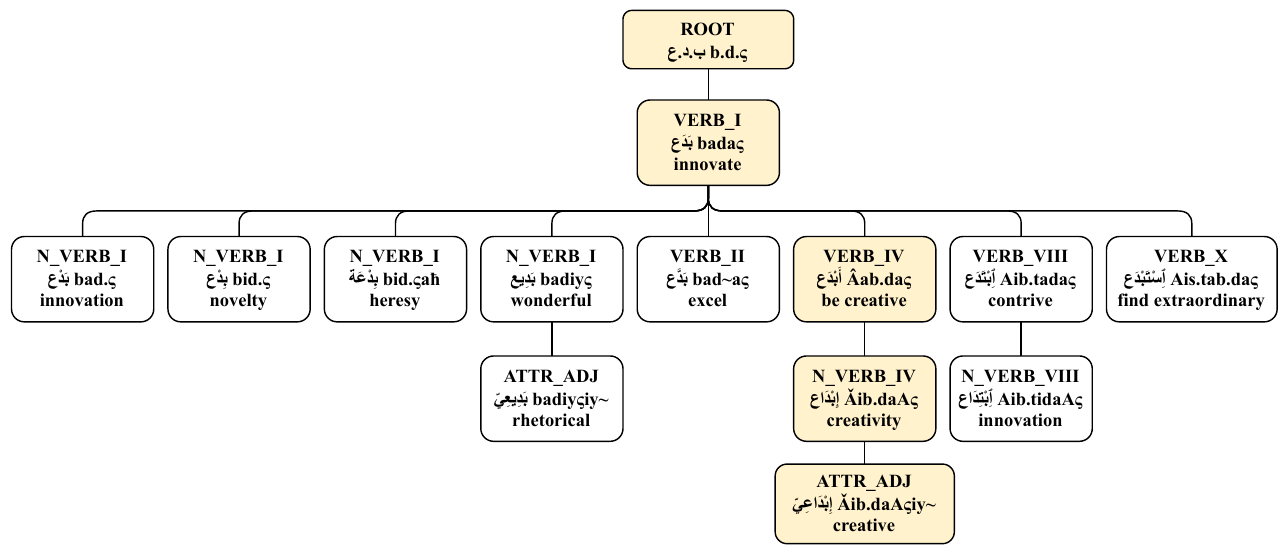}
    \caption{An example of a collection of derivative chains from one root. Highlighted is a chain that links a number of lemmas in derivational progression: the root \<ع>.\<د>.\<ب> \textit{b.d.{\AYN}} `innovation related'
    $\Rightarrow$ the verb \<بَدَع> \textit{bada{\AYN}} `to innovate'  
    $\Rightarrow$ the verb \<أبْدَع> \textit{{\AHAMZAUP}b.da{\AYN}} `to be creative'
    $\Rightarrow$ the noun \<إِبْدَاع> \textit{{\AHAMZADN}ibdaA{\AYN}} `creativity'
    $\Rightarrow$ the adjective \<إِبْدَاعِيّ> \textit{{\AHAMZADN}ibdaA{\AYN}iy{\SHADDA}} `creative'.
    }
    \label{fig:b.d.E}
     \end{center}
\end{figure*}
\subsection{The Abstract Level}

The  network we developed  covers all potential connections between roots and their derived patterns in a tree structure. The roots are positioned at the apex of the tree, followed by unaugmented verbs, and subsequently the augmented verbs along with the nominal derivatives. This network is meticulously organized to display all conceivable connections between patterns, even if certain connections may not be attested but remain theoretically plausible.
 
\paragraph{Constructing the Abstract Network} 
First, we classified all patterns according to their morphosemantic characteristics. Appendix~\ref{app:classes}~(Table~\ref{tab:classes}) presents examples of  the adopted classification of the patterns selected in this study. 

Next, we devised a scheme  to incorporate the derivational features of these patterns into the network.  The construction of the network involved the establishment of three tables to represent the source and target nodes, along with their relationships.


\begin{enumerate}

\item \textbf{The Canonical Table} 
We manually constructed a table that covers trilateral and quadrilateral verbs in  their unaugmented and augmented forms as well as all their derived nominal patterns.  We present examples of the Canonical Table entries in Appendix~\ref{app:canonic_table}: Table~\ref{tab:tablePI} focuses on forms connected to the triliteral unaugmented verbs (Form I,~\<فعل>~\textit{CaCaC/1a2a3}), and Table~\ref{tab:tablePARTII} includes the rest.


\item \textbf{{The Affixational Table}} 
To model affixational derivatives, we automatically generate new entries combining  Canonical Table entries with specific derivational affixes. For instance, the Canonical Table pattern \textit{1i23} is extended to \textit{1i23+iy{\SHADDA}} to allows us to model the example  \<علمي> \textit{{\AYN}ilmiy{\SHADDA}} `scientific' from \<علم> \textit{{\AYN}ilm} `science' discussed in Section~\ref{terms}.

\item \textbf{{The Semantic Specification Table}}
To model derivations that involve a semantic specification shift without a change in the main abstract pattern, we automatically generate entries from the Canonical and Affixational Tables, under a set of manually specified constraints. A major type of such entries involves (but not only) a change in part-of-speech; and in some cases, it may use an inflected form such as the feminine singular or broken plural. For instance,
the derivation \textit{ma12uw3+a{\TAMARBUTA}} (noun) from \textit{ma12uw3} (adjective) allows us to model the derivation of  \<معلومة> \textit{ma{\AYN}luwma{\TAMARBUTA}} `a piece of information (noun)' from the feminine form of \<معلوم> \textit{ma{\AYN}luwm} `known (adj)'. 

\end{enumerate}

The final Abstract Network is built as a combination of  the above-mentioned tables in one relational database to allow for efficient access to the chains of connected patterns, all their features, and their derivational relations.

\subsection{The Concrete Level}

Next we discuss aligning the {\camelmorph} database lemmas with the abstract network to create the {\chainbank}.

For each collection of lemmas from the same root, a derivational family, we recursively construct a tree starting with the root.
We only add children (derived lemmas) to parents (derivational bases or the root) in the tree if they match an allowable abstract network pair in terms of all linguistic attributes of child and parent.\footnote{In some cases, we require an inflectional process as an intermediary stage to produce a new derivation pattern, e.g., deriving a lemma pattern from the plural form of its base lemma: the  attributive adjective \<حدودي> {\it Huduwdiy{\SHADDA}} `bordering' is derived from the plural form of \<حد> {\it Had{\SHADDA}}  `border' (\<حدود> {\it Huduwd} `borders').}
If a lemma could be paired with different parents or with the same parent but with different relations we duplicate the child lemma and assign it as many times as needed.
We continue to assign children to parents until we exhaust all possible pairings. 
The result  is ideally a fully connected tree (knowledge graph) starting with the root of the derivational family and including chains that link it to every lemma in the family. In addition to the lemmas, the nodes of the tree include key linguistic attributes such as the part-of-speech and derivational class.  Figure~\ref{fig:b.d.E} is an example from the {\chainbank}.

In some cases, we may end up with disconnected subtrees due to a lack of allowed pairings in the abstract network. 
This may be the result of patterns that are disused with some roots.\footnote{A solution to consider in the future is to force attach such disconnected subtrees to the root with an \textit{Unknown} relation.}


\begin{table*}[th!]
\begin{center}
\begin{tabular}{|l|rr|rr|rr|}
\cline{2-7}
                    \multicolumn{1}{c}{}    & \multicolumn{2}{|c}{\textbf{Assessment 1: Dev}}    & \multicolumn{2}{|c}{\textbf{Assessment 2: Test}}   & \multicolumn{2}{|c|}{\textbf{All}}     \\ \hline
\textbf{Roots}          & 25 &              & 75        &        & 4,924      &        \\
\textbf{Lemmas}         & 566        &      & 1,608       &      & 34,453     &        \\
\textbf{Detected Relations} & 496 &(87.63\%) & 1,147 &(71.33\%)  & 23,333 &(67.72\%)  \\\hline
\textbf{Single Relation Correct} & 450 &(90.73\%) & 1,058 &(92.24\%) &        \multicolumn{1}{c}{}           \\
\textbf{Multiple Relation Correct} & 45 &(9.07\%)  & 76 &(6.63\%)    &    \multicolumn{1}{c}{}               \\
\textbf{No Correct Relation} & 1 &(0.20\%)   & 13 &(1.13\%)     &        \multicolumn{1}{c}{}         \\ \cline{1-5}
\end{tabular}
    
    \caption{
    Results of constructing the relational derivational {\chainbank} using the {\camelmorph} database, on development (Dev) and test subsets of the roots, and on all roots.}
\label{tab:results}
\end{center}
\end{table*}

\section{Evaluation}

\subsection{Experimental Setup}

\paragraph{Gold Chains} We manually constructed a set of 100 {\chainbank} trees correspondign to 100 randomly selected roots from {\camelmorph}.  To speed up the process, we started with automatically generated trees using an earlier version of our approach, and manually corrected them.

\paragraph{Data Splits} We split our 100 {\chainbank} trees into two sets: Dev (25 roots) and Test (75 roots). We used the Dev set to help debug and improve the tables we created for the abstract network and optimize our algorithm. The Test is used for reporting on the final implementation.

\paragraph{Metrics} we report on the following metrics.
\begin{itemize}
    \item Detected Relations (\%) is the percentage of all Gold Chain lemmas we detected automatically.
    \item Single Relation Correct (\%) is the percentage of all Gold Chain detected relations that unambiguous and correct.
    \item Multiple Relation Correct (\%) is the percentage of all Gold Chain detected relations that ambiguous and but with one correct answer.
    \item No Correct Relations (\%) is the percentage of remaining detected relations.
\end{itemize}

\subsection{Results and Discussion}

The results on Dev show a high degree of detected relations, but not perfect ($\sim$88\%), with over 90\% single relation correct.  The Test is lower in terms of detected relations ($\sim$71\%), but a higher single relation correct (92\%).
Multiple relations, accounting for $\sim$6-9\% of the cases in Dev and Test, occur due to shared patterns across different derivational classes. Missing relations stem from three main factors: (i) the relational data lacks primary nouns and other nominal lemmas, which require specific paths in the {\chainbank}s; (ii) {\camelmorph}’s database wasn't designed for derivational modeling, resulting in incomplete lemma groups for some roots and chain disconnections; or (iii) the relational database needs expansion with new non-canonical patterns. Additionally, the system could be improved by adding features and techniques to resolve ambiguities during evaluation.

All results are presented in Table~\ref{tab:results}.

\subsection{{\chainbank} v1.0}
We further applied our system on 4,926 roots from {\camelmorph} and their lemmas, which  resulted in 23,333 relations ($\sim$68\% detected relations), constituting the first version of the {\chainbank}. We plan to manually correct and further annotate additional entries in the future.
The Arabic Derivational {\chainbank}~v1.0
is publicly available to support further Arabic NLP research.\footnote{\url{https://github.com/CAMeL-Lab/ArabicChainBank}} 


\section{Conclusion and Future Work}

 We introduced the Arabic Derivational {\chainbank} framework for modeling Arabic derivational morphology. The evaluation of our rule-based method to populate the {\chainbank} shows great promise. The first edition of the {\chainbank}  and its  framework are publicly available.

%

Future work will continue to expand the abstract network to include missing patterns, including  non-canonical patterns, and to develop advanced disambiguation techniques to further enhance the {\chainbank}. 
This work should happen in tandem with improving the coverage of {\camelmorph} in terms of lemmas and their features.
Our long term vision is to include dialectal lemmas in a manner that shows their connections with each other and with Standard Arabic lemmas.


\section{Limitations}

We acknowledge several limitations of the work as presented. First, the reliance on a rule-based methodology, although efficient, may overlook nuances that a more comprehensive manual annotation process could capture. This could lead to the omission of certain derivational patterns and relations. Second, the alignment with the {\camelmorph} morphological analyzer, though beneficial for broad coverage, may have resulted in incomplete or fragmented derivational chains due to the database’s current structure, which was not designed for derivational modeling. Third, the dataset predominantly covers canonical derivational patterns, with non-canonical patterns remaining underrepresented, potentially limiting the \chainbank’s applicability to broader linguistic phenomena.  Lastly, the work focuses on 
Standard Arabic and does not cover any of its major dialects.  Future work will address these limitations to enhance the framework's completeness and accuracy.

\bibliography{custom,anthology,camel-bib-v3}

\onecolumn
\newpage
\appendix
\section{{\chainbank} Derivational Classes}
\label{app:classes}

\begin{table*}[h!]
\begin{center}
 \includegraphics[width=0.9\textwidth]{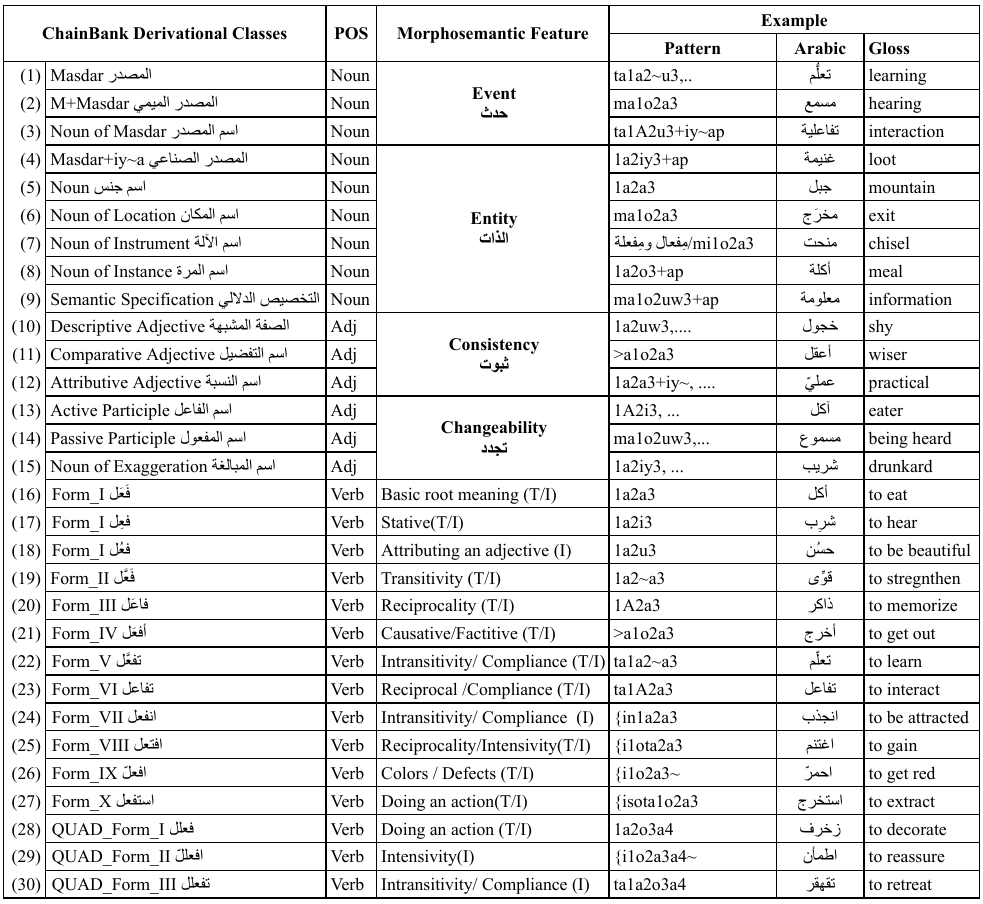}
    \caption{{\chainbank} Derivational Classes: an overview of Arabic Morphosemantic patterns with examples. (T/I) refers to transitive \& intransitive}
\label{tab:classes}
\end{center}
\end{table*}

\newpage
\section{{\chainbank} Tags}
\label{app:tags}

\begin{table*}[h!]
%
\begin{center}
 \includegraphics[width=\textwidth]{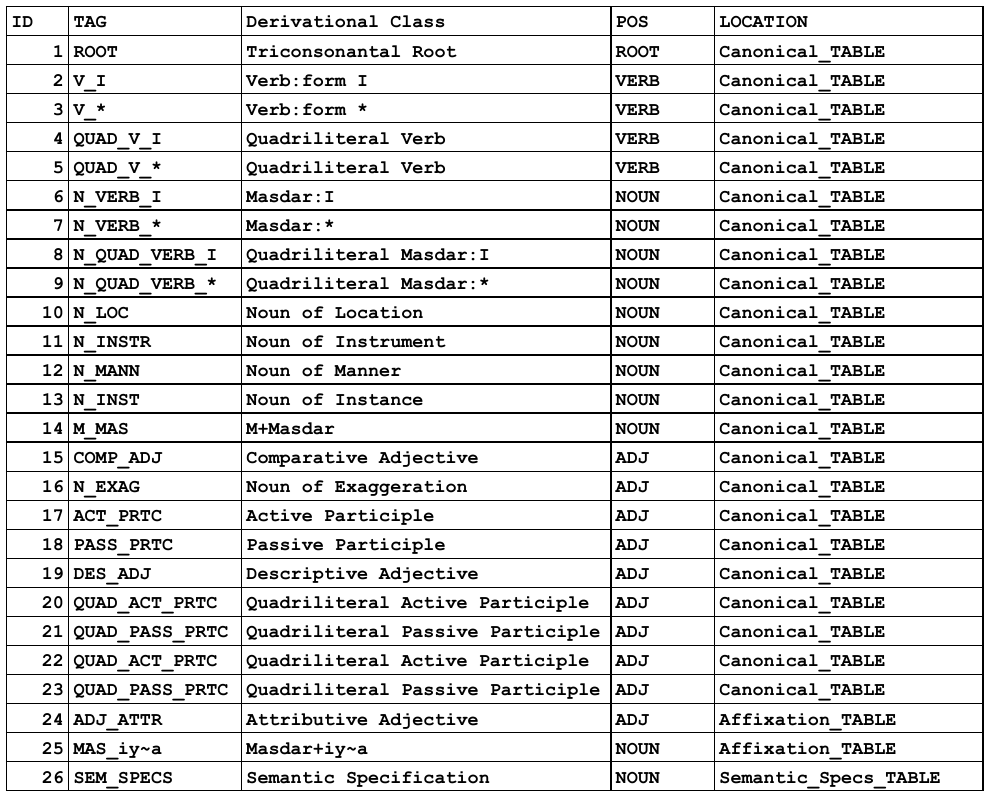}
 
\caption{Derivational Classes in the Arabic {\chainbank}, showing ID, tag, class,  POS, and location of each form within the derivational network.}
\label{tab:tags}   
\end{center}
\end{table*}


\newpage
\section{{\chainbank} Canonical Tables}
\label{app:canonic_table}

\begin{table*}[ht!]
\begin{center}
 \includegraphics[width=\textwidth]{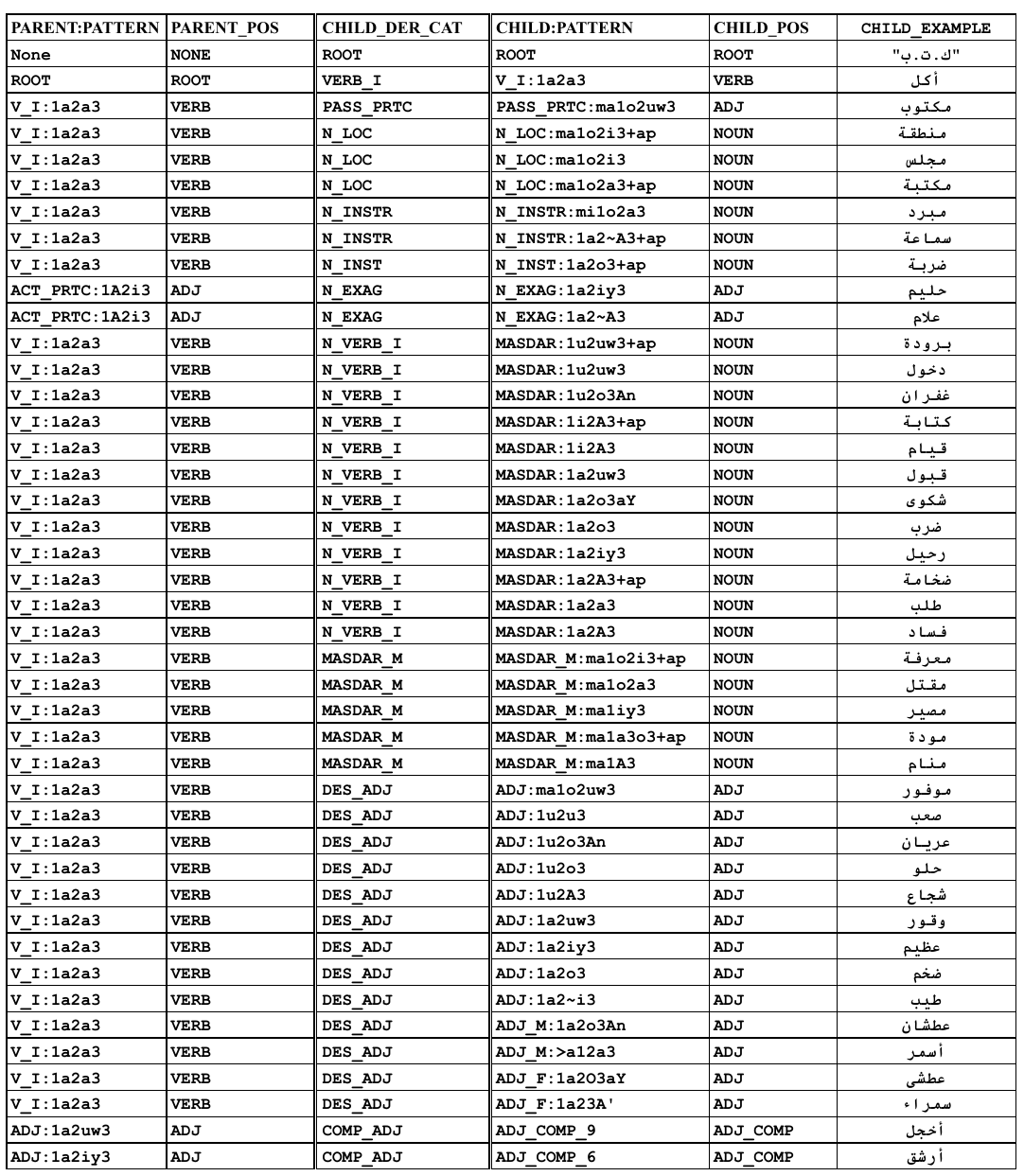}
 
\caption{(Part I) A sample of entries from \texttt{Canonical\_I} table: one of the fundamental tables in the {\chainbank}.}

\label{tab:tablePI}
\end{center}
\end{table*}

\begin{table*}[ht!]
\begin{center}
 \includegraphics[width=\textwidth]{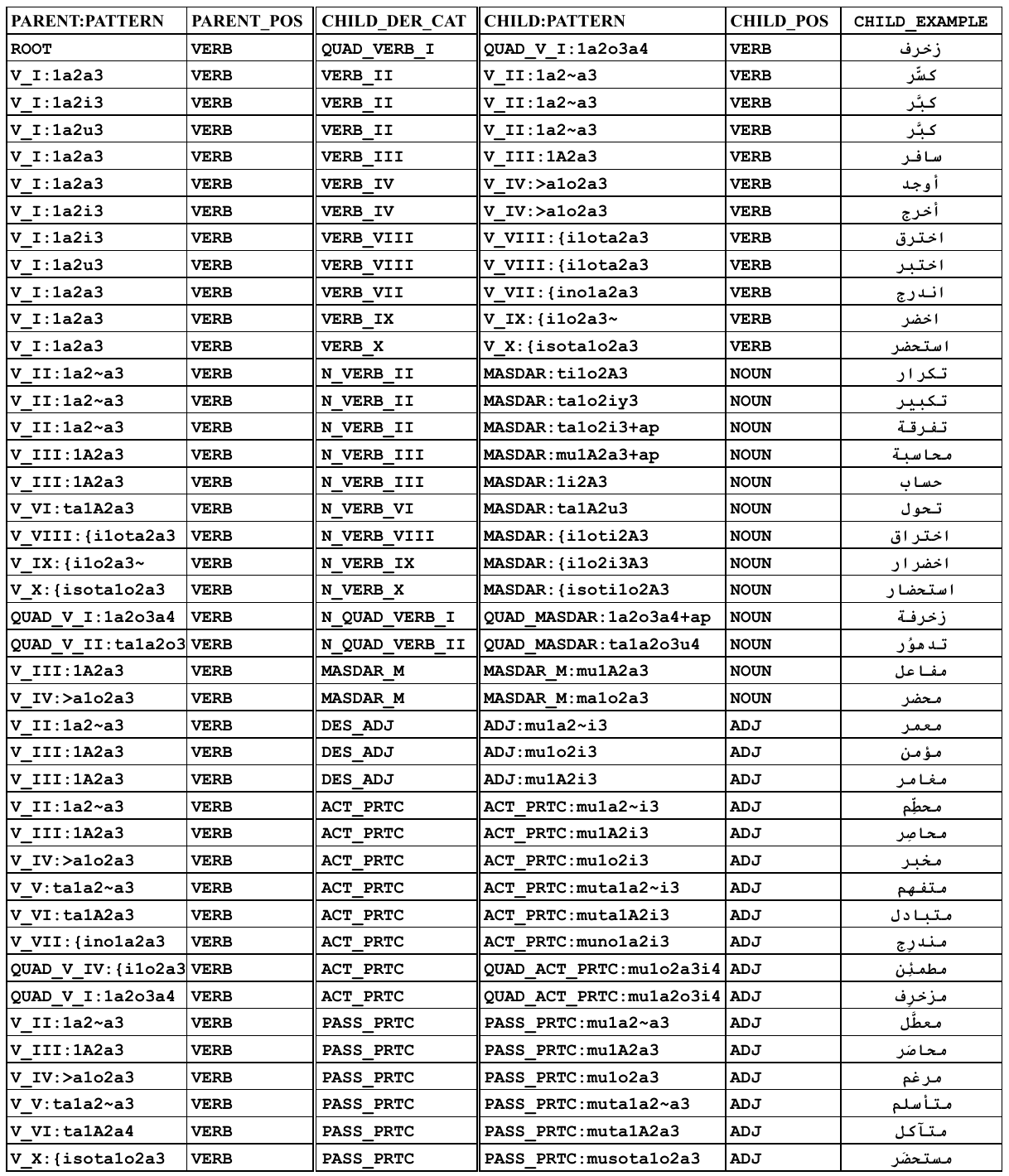}
 
\caption{(Part II) A sample of entries from \texttt{Canonical\_*} table: one of the fundamental tables in the {\chainbank}.}
\label{tab:tablePARTII}
\end{center}
\end{table*}

\end{document}